\documentclass[fleqn,10pt]{wlscirep}
\usepackage{graphicx}
\usepackage{subcaption}
\usepackage{array}
\usepackage{float}
\usepackage{amsmath}
\usepackage{amssymb}
\usepackage{algorithm}
\usepackage[noend]{algpseudocode}

\newcommand{\norm}[1]{\left\lVert#1\right\rVert}

\title{Variational Encoding of Complex Dynamics}

\author[1]{Carlos X. Hern\'{a}ndez \thanks{These authors contributed equally to this work}}
\author[2]{Hannah K. Wayment-Steele $^{*}$}
\author[2]{Mohammad M. Sultan $^{*}$}
\author[2]{Brooke E. Husic}
\author[1,2]{Vijay S. Pande \thanks{Corresponding author: pande@stanford.edu}}
\affil[1]{Biophysics Program, Stanford University, California, USA}
\affil[2]{Chemistry Department, Stanford University, California, USA}

\begin{abstract}
Often the analysis of time-dependent chemical and biophysical systems produces high-dimensional time-series data for which it can be difficult to interpret which individual features are most salient. While recent work from our group and others has demonstrated the utility of time-lagged covariate models to study such systems, linearity assumptions can limit the compression of inherently nonlinear dynamics into just a few characteristic components. Recent work in the field of deep learning has led to the development of variational autoencoders (VAE), which are able to compress complex datasets into simpler manifolds. We present the use of a time-lagged VAE, or variational dynamics encoder (VDE), to reduce complex, nonlinear processes to a single embedding with high fidelity to the underlying dynamics. We demonstrate how the VDE is able to capture nontrivial dynamics in a variety of examples, including Brownian dynamics and atomistic protein folding. Additionally, we demonstrate a method for analyzing the VDE model, inspired by saliency mapping, to determine what features are selected by the VDE model to describe dynamics. The VDE presents an important step in applying techniques from deep learning to more accurately model and interpret complex biophysics.
\end{abstract}

\begin{document}
\flushbottom
\maketitle
\thispagestyle{empty} 

\section{Introduction}
Simulations of biomolecules have provided insight into molecular processes with increasing time- and length-scales due to advances in both algorithms \cite{shirts2000screen} and hardware \cite{shaw2008anton}. Such simulations can have thousands of degrees of freedom, making it crucial to have meaningful and statistically robust methods to extract underlying dynamical processes \cite{Shukla2015}. 

The dynamics of molecular systems are often represented using the dynamical propagator approach \cite{prinz2011markov}. Given an ensemble of particles at time $t$ distributed in phase space with a given probability distribution $p(x,t)$, we seek to describe a propagator, an operator that can describe the new distribution of the ensemble, $p(x,t+\tau)$, given some lag time $\tau$. When $\tau$ is chosen such that these probabilities are independent of the history of the system, the model is said to be Markovian.  Many methods have been developed to compute approximations to the propagator of a molecular system from simulation data, including time-structure-based independent component analysis (tICA)\cite{Naritomi2011, Perez-Hernandez2013, Schwantes2013} and extensions (kernel tICA \cite{Schwantes2015}, sparse tICA \cite{McGibbon2015}), Markov State models (MSM) \cite{bowman2013introduction}, VAMPnets \cite{wu2017variational}, soft-max MSMs \cite{harrigan2017}, and diffusion maps \cite{coifman2006diffusion}.

\begin{figure}[H]
    \centering
    \includegraphics[width=.4\textwidth]{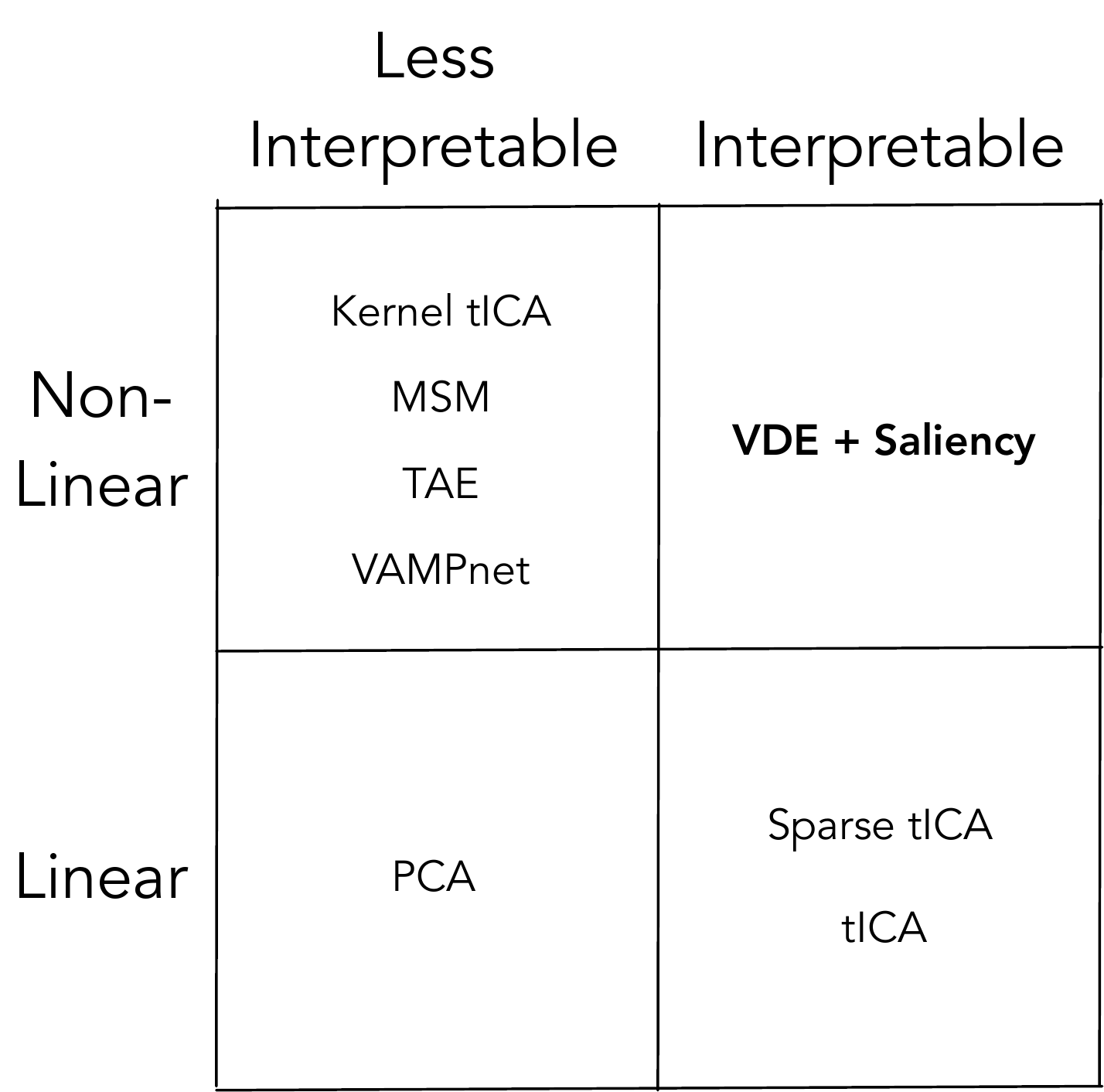}
    \caption{An overview of a subset of the methods used to analyze protein dynamics in terms of model interpretability and ability to capture non-linear motions \cite{Pearson1901, Naritomi2011, Perez-Hernandez2013, Schwantes2013, Schwantes2015, harrigan2017landmark, McGibbon2017, Mardt2017, Wehmeyer2017}. Here, we define interpretability as the ease with which the scientist can analyze the model for feature importance with respect to dynamics. For example, principal component analysis (PCA), arguably the simplest model mentioned, is typically ill-suited to analyze complex dynamics and, therefore, principal components are not reliably meaningful. Meanwhile, the VDE is able to leverage deep learning to model non-linear relationships between time-dependent observables and saliency mapping to understand which observables contribute most to the model.}
    \label{fig:methods}
\end{figure}

Any method for approximating the dynamics of a complex system has two objectives: to adequately represent complexity in the form of model nonlinearity and to be interpretable, that is, to be readily analyzable for feature importance. In Figure \ref{fig:methods}, we indicate how several commonly-used methods for dimensionality reduction of dynamical systems compare in terms of achieving these two aims. Complexity and interpretability often come at the expense of each other. For instance, kernel methods such as kernel tICA \cite{Schwantes2015, harrigan2017landmark} improve the ability to capture nonlinear effects of features in dynamics over linear methods; however, identifying biophysical meaning in coordinates in an implicit kernel space remains a challenge. Conversely, standard tICA and sparse tICA allow for more precisely identifying relevant biophysical features, but the linearity of tICA limits the complexity of dynamics it can represent.

An alternative technique for dimensionality reduction is the autoencoder framework \cite{Kingma2013, rezende2014stochastic}. An autoencoder is a deep unsupervised learning algorithm that aims to learn a low-dimensional representation of high-dimensional data \cite{Hecht-Nielsen1995, Hinton2006}. An autoencoder has two components: an encoder network and a decoder network. The encoder network reduces the input data to a low-dimensional representation, referred to as the latent space of the autoencoder, and the decoder network reconstructs the latent representation to the original dimensionality. The difference between the original data and the reconstruction is used to update and train the network. Variational autoencoders (VAEs) add regularization to the encoder framework by applying Gaussian noise to the latent space \cite{Bengio2013}. The term ``variational'' stems from this stochasticity: the autoencoder is an implementation of variational Bayesian inference with a Gaussian prior, which maximizes the lower bound on the log-likelihood of the observed data \cite{Kingma2013}.

Recently, the autoencoder framework has been extended to model time-series data \cite{Walker2016, grathwohl2016disentangling, fraccaro2017disentangled, bao2017deep, cui2017deep, doerr2017dimensionality, Wehmeyer2017}. Analysis in these applications typically involves mapping time-series data to latent spaces with the same dimensionality as the length of the initial time-series data and has not focused on approximating a propagator for the time-series data; however, there are a couple of notable exceptions. Doerr and De Fabritis \cite{doerr2017dimensionality} recently compared a simple autoencoder to other methods for dimensionality reduction of biophysical simulation data. Wehmeyer and Noe introduced a time lag into an autoencoder (TAE) framework to describe dynamics~\cite{Wehmeyer2017}. Interestingly, they demonstrate that in the limit of a single linear hidden layer, the tICA solution can be attained.

In this work, we extend the traditional VAE architecture to approximate a propagator for time-series data in an architecture denoted as a Variational Dynamics Encoder (VDE). This represents the first use of a time lag within a variational autoencoder to our knowledge. Additionally, we introduce a novel ``autocorrelation'' loss function, which is inspired by the variational approach to conformational dynamics (VAC) \cite{Noe2013}. We demonstrate that this approach yields models with more explanatory power than linear dimensionality reduction techniques in both the M\"uller-Brown potential and the folding landscape of the villin headpiece subdomain.
We also explore the generative capability of the VDE as a propagator of dynamics and show that, as implemented, it is unable to reliably capture thermodynamics at differing temperatures.
Finally, we demonstrate a novel analysis method, inspired by saliency mapping in neural nets for visual classification~\cite{Simonyan2013, Springenberg2014,Saliency2013}, to lend interpretation to VDE models. This combination of using the VDE with saliency mapping creates a framework that enables nonlinear combinations of features while remaining interpretable.

\section{Model: Variational Dynamics Encoder (VDE)}
\subsection{VDE Architecture}
\label{sec:arch}

\begin{figure}[H]
    \centering
    \includegraphics[width=.5\textwidth]{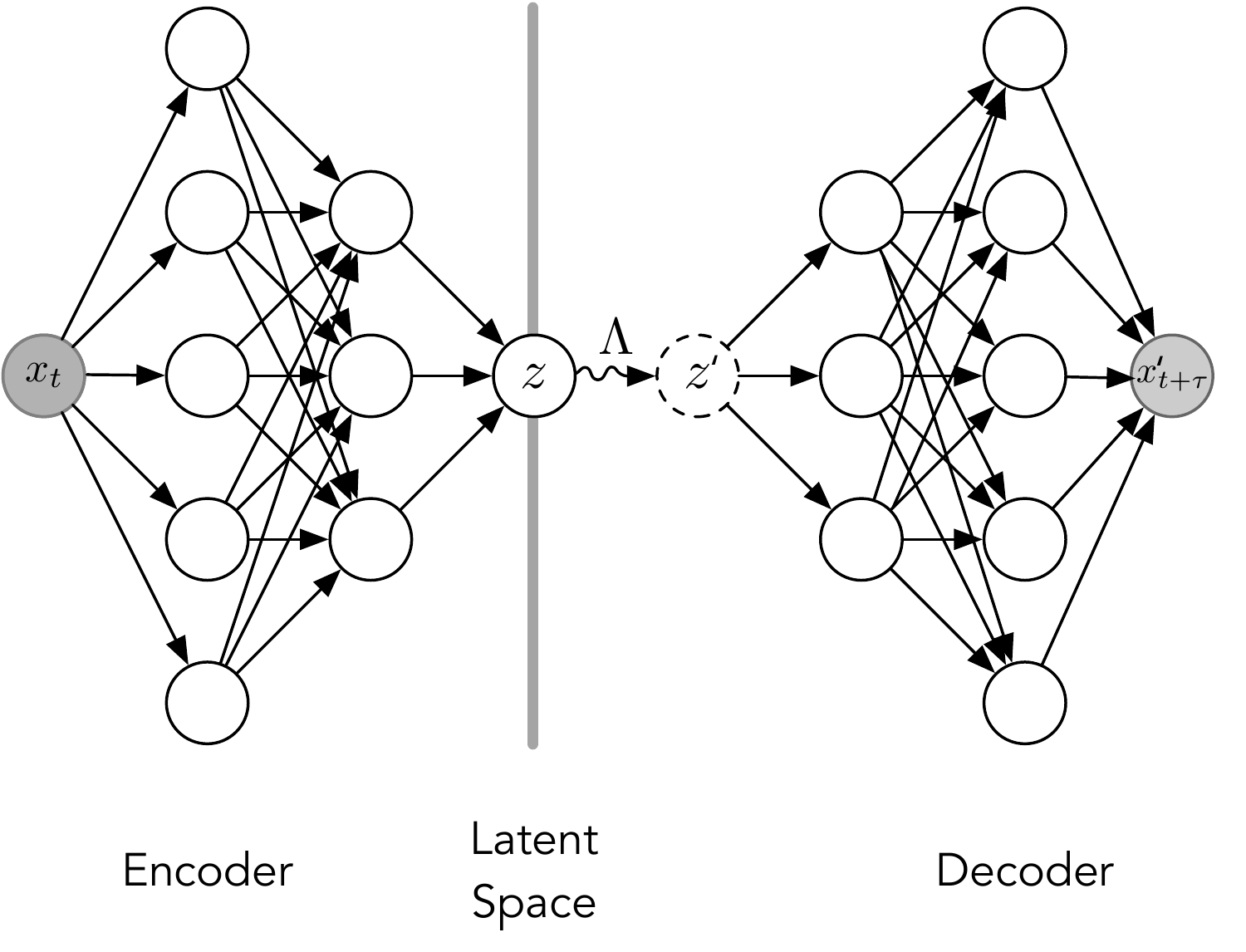}
    \caption{A schematic of the VDE. Features, $x_t$, at some timepoint, $t$, are fed into the network in order to make a prediction of the state of the system, $x^{\prime}_{t+\tau}$, at a future timepoint, $t + \tau$, where $\tau$ is some Markovian lag time. As with a traditional VAE, the network can be
    subdivided into three parts:~the encoder network; variational layer, $\Lambda$; and the decoder network, as labeled. Our encoder network is a DNN with non-linear activation functions in the hidden layers, which eventually bottlenecks into the one-dimensional latent space, $z_t$. The latent space is then slightly perturbed with Gaussian noise by the $\Lambda$-layer to generate $z^\prime$, as described by Kingma and Welling~\cite{Kingma2013}. Finally, the decoder network, also a DNN, mirrors the encoder network in architecture by using $z^\prime$ to generate $x^{\prime}_{t+\tau}$, a prediction of how the system will evolve after one lag time of $\tau$.}
    \label{fig:network}
\end{figure}

The architecture of the VDE, as seen in Figure \ref{fig:network}, closely resembles that of a VAE; however, the training procedure is slightly modified to suit time-series data \cite{doerr2017dimensionality, Wehmeyer2017}. The most significant modification being that featurized data, $x_t$, at some timepoint, $t$, are fed into the network in order to make a prediction of the state of the system, $x^{\prime}_{t+\tau}$, at a future timepoint, $t + \tau$, where $\tau$ is some user-selected lag time such that the dynamics of the system is Markovian. As with a traditional VAE, the network can be subdivided into three parts: the encoder network; variational layer, $\Lambda$; and the decoder network.

The encoder network is a deep neural network (DNN) with non-linear activation functions and a user-selected number of hidden layers, which eventually bottlenecks into the one-dimensional latent space, $z_t$. In this way, the encoder network functions as a non-linear dimensionality reduction of $x_t$. The latent space is then perturbed by Gaussian noise within the $\Lambda$-layer, with mean parameter, $\mu$, variance parameter, $\sigma^2$, and arbitrary scaling, $\alpha$, to generate $z^\prime$, as described by Kingma and Welling~\cite{Kingma2013}. Finally, the decoder network, also a DNN, mirrors the encoder network in architecture by using $z^\prime$ to generate $x^{\prime}_{t+\tau}$, a prediction of how the system will evolve after a duration of $\tau$.

Once the VDE has been trained, it can be used for both dimensionality reduction and synthetic trajectory generation.
During dimensionality reduction, only the encoder network is necessary, which provides a direct mapping of $x \mapsto z$.
During trajectory generation, the entire VDE network is needed. An initial set of features, $x_0$, is fed through the network to generate $x^{\prime}_\tau$, the predicted state after a duration of $\tau$. This can be done iteratively to generate an arbitrarily long trajectory of features exhibiting thermodynamics consistent with that of the original system used during training. In over to overcome the model's insensitivity to $\Lambda$-layer used during training, we recommend that the noise scaling, $\alpha$, is increased such that $\alpha_{\text{generation}} \gg \alpha_{\text{train}}$.

\subsection{VDE Loss Function}
\label{sec:loss}

The VDE is quantitatively evaluated by calculating the sum of two loss functions: reconstruction loss and autocorrelation loss:
\begin{equation}
\mathcal{L}_\text{VDE} = \mathcal{L}_\text{reconstruction} + \mathcal{L}_\text{autocorrelation}.
\end{equation}
The first of these two, reconstruction loss, attempts to quantify how well the VDE approximates the state of the system at $t+\tau$, given the true state of the system at time $t$ \cite{Bengio2013, Kingma2013}. In doing so, we are essentially evalutating the ability of the latent space to approximate the Markovian propagator after a single lag time. This can be done by considering the mean squared error between the predicted propagation, $x^{\prime}_{t+\tau}$, and the true propagation, $x_{t+\tau}$, along with the Kullback–Leibler divergence between latent space priors that generate $x^{\prime}_{t+\tau}$:
\begin{equation}
\mathcal{L}_\text{reconstruction} = \norm{x^\prime_{t+\tau} - x_{t+\tau}} + \frac{1 + \log\sigma^2 - \mu^2 - \sigma^2}{2},
\end{equation}
\noindent{}where $\mu$ and $\sigma$ are learnable parameters representing the mean and standard deviation of the Gaussian prior applied to the latent space, respectively. Together, these complementary loss terms capture a trade-off between model complexity and simplicity of the Gaussian prior. Reconstruction loss pushes the model towards having high fidelity to the training data, while the Kullback–Leibler divergence acts as a regularization term to ensure that the latent space behaves as a Gaussian emission~\cite{Kingma2013}.

The autocorrelation loss attempts to optimize the network towards a more complete representation of the long time-scale kinetics observed within time-series. Although minimization of the reconstruction loss has the potential to recover these dynamical processes~\cite{Mardt2017}, we find that in some cases, such as in Section \ref{sec:villin}, it alone is not sufficient.
In order to improve model convergence, we borrow from the VAC~\cite{Noe2013}, a specific application of the variational principle from quantum mechanics adapted for Markov modeling and a useful tool for parameter selection. The variational principle states that, in the limit of infinite data, no process can be identified in the data that is slower than the true process. If we interpret the variational principle as the measure of the quality of this approximation, the phase-space decomposition that leads to a linear model with larger leading dynamical eigenvalues is consequently the better phase-space decomposition~\cite{McGibbon2015}. In the limit of a single linear decomposition of phase-space, there is only one eigenvalue to consider, which is equivalent to the autocorrelation of $z$ \cite{Noe2013}:
\begin{equation}
\mathcal{L}_\text{autocorrelation} =  - \rho_{z_t, z_{t+\tau}} = - \frac{\mathop{\mathbb{E}}\left[ \left(z_t - \bar{z}_t \right) \left( z_{t+\tau} - \bar{z}_{t+\tau} \right) \right]}{s_{z_t} s_{z_{t+\tau}}}, 
\end{equation}
where $\bar{z}$ and $s_z$ are the sample mean and population mean of the latent space for a particular batch of data, respectively. For linear models, this leads only to a first-order approximation of slowest process; however, by incorporating this into the VDE's loss function, we take advantage of the deep encoder as a general approximator to the slowest processess found within in our data~\cite{noe2011dynamical}. Algorithm \ref{train} outlines how these two losses are calculated and used for backpropagation in practice. Note that the data is split into many smaller batches during training, with $x_t$ as input variable and $x_{t+\tau}$ as the target variable, to take advantage of stochastic gradient descent methods.  We also recommend pre-processing features—either via standardization or median-centering and scaling by interquartile ranges—to prevent the reconstruction loss from overpowering the autocorrelation loss~\cite{msmbuilder}.

\begin{algorithm}[H]
\caption{Training the VDE}\label{train}
  \begin{algorithmic}[1]
    \Procedure{Train}{\texttt{model}, \texttt{data}}
      \For{\texttt{batch} $\in$ \texttt{data}}
        \State $x_t$, $x_{t+\tau} \gets$ batch
        \State $z_t \gets$  \texttt{model.encode($x_t$)} 
        \State $z^\prime_t \gets$ \texttt{model.lambda($z_t$)} 
        \State $x^\prime_{t+\tau} \gets$ \texttt{model.decode($z^\prime_t$)}
        \State $z_{t+\tau} \gets$ \texttt{model.encode($x_{t+\tau}$)}
        \State
        \State $\mu, \sigma \gets$ \texttt{model.lambda.parameters} 
        \State
        \State $\mathcal{L}_\text{reconstruction} \gets \norm{x^\prime_{t+\tau} - x_{t+\tau}} + \frac{1 + \log\sigma^2 - \mu^2 - \sigma^2}{2}$ 
        \State $\mathcal{L}_\text{autocorrelation} \gets - \rho_{z_t, z_{t+\tau}}$
        \State
        \State \texttt{model.loss} $\gets \mathcal{L}_\text{reconstruction} + \mathcal{L}_\text{autocorrelation}$
        \State
        \State \texttt{model.loss.backward()}
        \State \texttt{model.optimizer.step()}
      \EndFor
    \EndProcedure
  \end{algorithmic}
\end{algorithm}

\section{Results}

\subsection{A Non-Linear Encoding for Brownian Dynamics}

\begin{figure}[H]
    \centering
    \includegraphics[width=.85\textwidth]{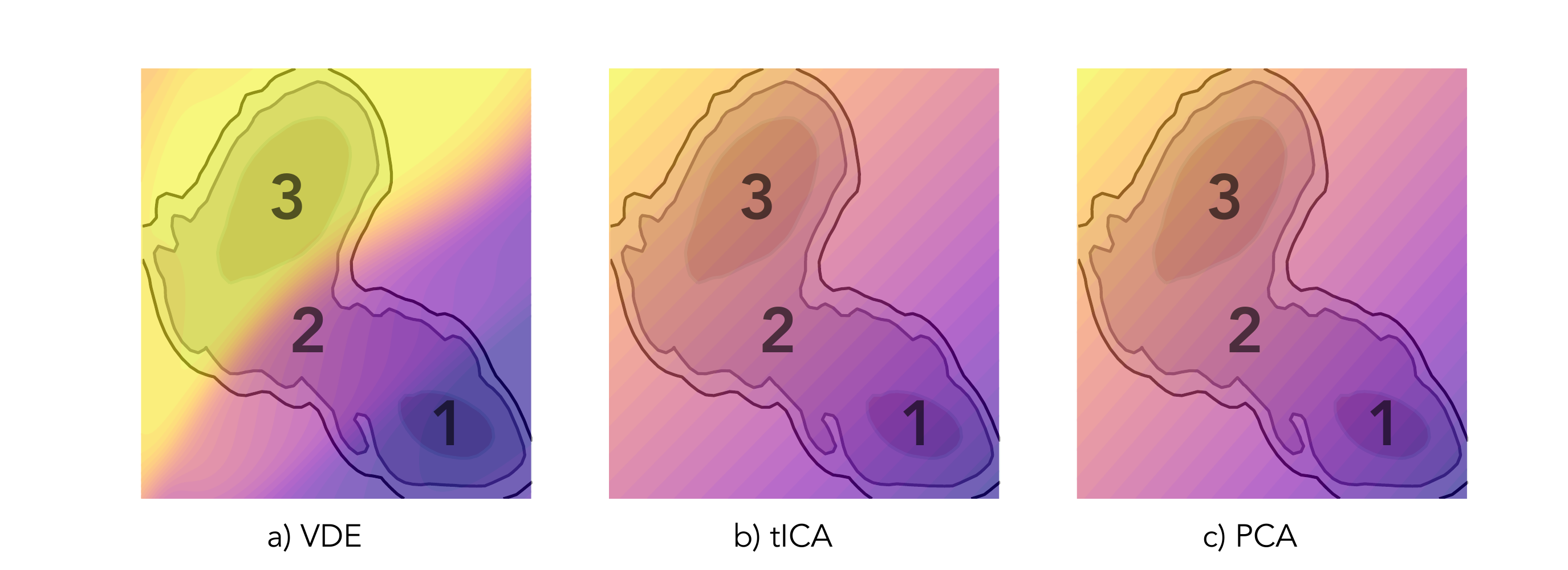}
    \caption{The 2-D M\"uller-Brown potential (gray-scale contours) overlaid with colormap projections of the one-dimensional a) VDE, b) tICA, and c) PCA coordinates. While tICA and PCA identify a strictly linear mode that approximates the slowest dynamical process (i.e. diffusion from region 1 to 3), the non-linear VDE is better able to map out basins (regions 1 and 3) and intermediate state (region 2). Note that because the region outside of the contours is energetically unfavorable, the color projections in that space are extrapolations of each method, respectively.
    }
    \label{fig:muller}
\end{figure}

We first apply the VDE framework to the well-studied 2-D M\"uller-Brown potential and demonstrate it can adequately describe the dynamics of this simple system. Figure \ref{fig:muller} shows results for a) the VDE, b) tICA, and c) PCA. We note that while tICA and PCA both identify the same dominant linear coordinate, representing diffusion from minor to major basin, the VDE generates a non-linear projection that is able to distiguish these basins more clearly, as well as the transition region. 

To establish an unbiased assessment of the VDE's performance compared to tICA or PCA, we constructed MSMs using identical hyperparameters and compared generalized matrix Raleigh quotients (GMRQs) of the slowest process as a scoring metric~\cite{McGibbon2015}. The VDE achieves a slightly higher mean GMRQ score ($1.8580\pm5 \times 10^{-4}$) than tICA ($1.8460\pm5 \times 10^{-4}$) or PCA ($1.8472\pm5 \times 10^{-4}$) on held-out data, suggesting that it is better able to represent the slowest timescale of this system.

\subsection{The VDE Does Not Behave as a True Propagator}

\begin{figure}[H]
    \centering
    \includegraphics[width=.85\textwidth]{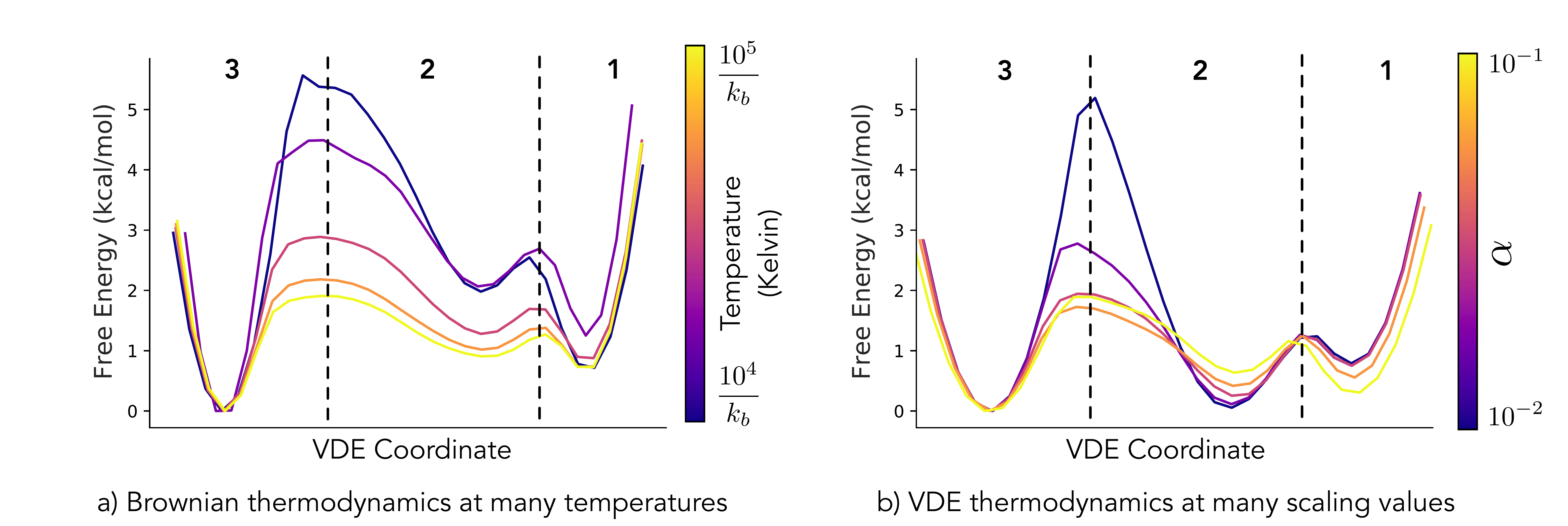}
    \caption{One-dimensional free energy projections generated from the VDE coordinate for a) true Brownian-dynamics simulations at different temperatures and b) fake trajectories generated by the VDE, trained at $k_{B}T = 1.5 \times 10^{4}$ Joules, with different scaling values, $\alpha$, a proxy for temperature, within the $\Lambda$-layer. Although not a true one-to-one comparison, we find that free energy barriers (between regions 1, 2, and 3) are lowered, as expected, when temperature is increased within the M\"uller-Brown potential; however, free energies of transition (region 2) and boundary regions (beyond regions 1 or 3) of phase-space cannot be reproduced reliably. We note that the selected $\alpha$ values have not been rigorously fitted to best match the different values of $k_{B}T$ shown, but were instead evenly sampled over a fixed interval, in which similar thermodynamics to simulation are observed.}
    \label{fig:propagator}
\end{figure}

As VAEs are regarded as a generative model, we consider the relationship between the VDE and the propagator function. When trained on the M\"uller-Brown potential, with $k_{B}T = 1.5 \times 10^{4}$ Joules, the VDE is able to generate ``fake'' trajectories with some similarities in thermodynamics to the original simulations, as seen in Figure \ref{fig:propagator} (pink curve). Furthermore, when we modulate the effect of the $\Lambda$-layer by adjusting the scaling parameter $\alpha$, we are also able to mimic some effects of changing the simulation temperature without having to re-train the VDE. Figure \ref{fig:propagator}b demonstrates that when decreasing (dark blue and purple curves) or increasing (orange and yellow curves) $\alpha$, the VDE is able to adjust barrier heights in a similar fashion to what is observed in simulation, shown in Figure \ref{fig:propagator}a. However, we find that this fidelity to simulation is lacking in transition regions and previously unobserved regions of phase-space, where the VDE does a poor job of recapitulating true thermodynamics.

Also of note is the case of $\alpha = 0$ (not shown in Figure \ref{fig:propagator}b), where the VDE behaves essentially as an indicator function, reporting which basin a given frame will eventually diffuse towards in a low temperature simulation. As $\alpha$ is increased, the VDE must decide which basin to push the now heat-bathed system towards and more realistic dynamics can be observed. Such behavior may be interpreted as a temperature-dependent propagator, whereby the VDE has learned some of the underlying thermodynamic characteristics of the system; although, there seems to be a strong attraction to certain basins (e.g. region 2) which is not observed in simulation. Because of this attraction, increasing $\alpha$ leads to the raising of intermediate basins towards realistic free energies by the VDE rather than the lowering one might expect when raising the temperature of a simulation. We recommend greatly increasing $\alpha$, as described in Section \ref{sec:arch}, when generating synthetic trajectories due to this trend.

\subsection{A Simple Encoding for Villin Headpiece Dynamics}
\label{sec:villin}

We next apply the VDE to pairwise alpha-carbon (C$\alpha$) contacts in order to model the folding process of the villin headpiece subdomain. Here, we aim to assess the quality of the VDE as a dimensionality reduction technique for protein folding by quantifying how well a MSM constructed from VDE-transformed data separates relevant timescales and distinguishes basins within the landscape. With these metrics in mind, the VDE appears to represent the folding landscape well and can even out-perform tICA using similar MSM hyperparameters.

\begin{figure}[H]
    \centering
    \begin{subfigure}[t]{0.8\textwidth}
        \centering
        \includegraphics[width=\textwidth]{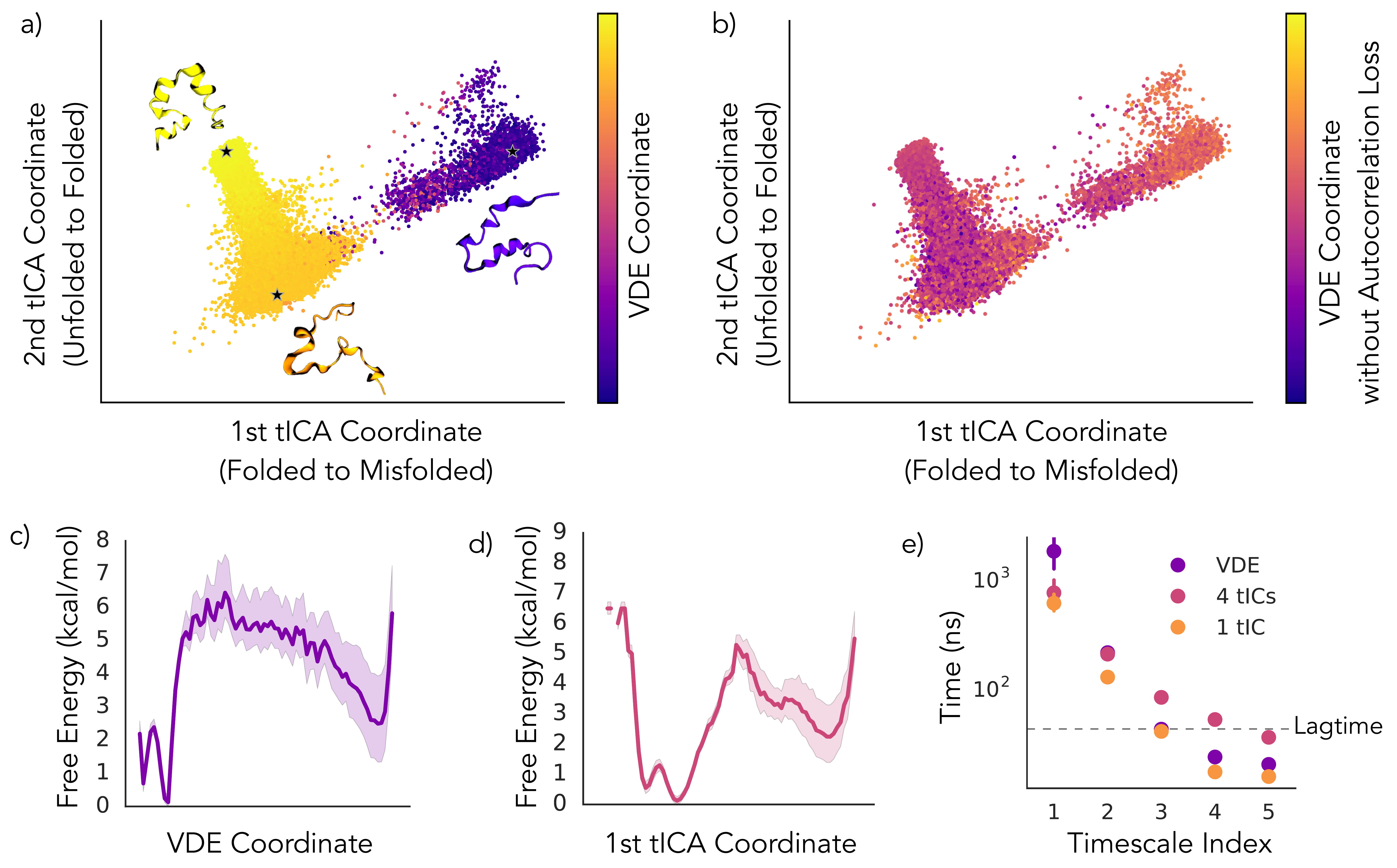}
    \end{subfigure}
    \caption{Folding coordinate for villin approximated by the VDE in comparison with the first two tICA solutions. a) Villin trajectory data projected onto the first two tICs of an optimized tICA model in order to compare the VDE coordinate to the two slowest coordinates from tICA, colored by VDE latent projection. Inset yellow, orange, and purple structures depict representative folded, unfolded, and misfolded structures, respectively.
    b) Villin trajectory data projected onto the first two tICs of optimal tICA model, colored by VDE latent projection without using the autocorrelation loss.
    c) Free energy of VDE latent projection. d) Free energy of first tICA coordinate. e) Comparison of timescales of MSMs constructed with a VDE model, optimized tICA model with 4 tICs, and tICA model with 1 tIC. In c), d), and e), error bars represent the range of 100 bootstrapped replicates. Where error bars are not visible, the error is too small to see.}
    \label{fig:villin}
\end{figure}

Figure \ref{fig:villin}a depicts trajectory data projected onto the slowest two tICA processes (tICs) from an optimized tICA model \cite{Husic2016} and colored by the projection onto the VDE latent coordinate. In the optimal tICA model, 2 tICs are needed to capture both the folding and a prominent misfolding process. The first tIC is unable to completely separate unfolded and folded state, whereas the second tIC distinguishes the folded and unfolded state but is unable to distinguish the folded and misfolded state. In contrast, the VDE latent coordinate is able to discriminate between all three states: folded, unfolded, and misfolded. By comparing the free energies of the VDE latent space (Figure \ref{fig:villin}c) and the first tIC (Figure \ref{fig:villin}d), we observe that the VDE coordinate has a narrower basin of folding than that of the first tIC, indicating the VDE latent coordinate more sharply resolves the folding basin than the first tIC does. 

To test the benefit of using the autocorrelation loss discussed in Section \ref{sec:loss}, we trained models of villin using only the reconstruction loss and no autocorrelation loss. The projection of the optimal model with no autocorrelation loss is portrayed in Figure \ref{fig:villin}b. In this projection, there is minimal differentiation between different parts of the landscape. This highlights the necessity of incorporating an autocorrelation loss into the VDE loss function.

MSMs for the villin landscape were constructed from both the VDE model and the optimized tICA model. Comparing these models indicates that the VDE model identifies a slower timescale than the tICA model. Figure \ref{fig:villin}e portrays the timescales of the slowest five processes identified by MSMs built from the VDE projection, our optimized tICA model, and a tICA model built with one tIC component. The timescale of the slowest process in the MSM from the VDE projection is $1620 \pm 80$ nanoseconds, whereas the timescale of the slowest process in the optimal tICA model is $770 \pm 40$ nanoseconds. According to the variational approach to conformational dynamics, as described in Section \ref{sec:loss}, a model with longer timescales should be closer to modeling the true dynamics of the system.

\subsection{Protein Saliency Maps Enable Interpretation of the VDE}
As noted in Figure \ref{fig:methods}, nonlinear methods for time-series analysis tend to sacrifice model interpretability.
Linear tICA provides ``loadings'' on each input feature for each slow mode. Thus, the absolute magnitude of these loadings can be used to understand holistic protein dynamics at the atomic scale\cite{msmbuilder}. To make VDEs more interpretable, we designed a novel variant to saliency maps (see Section \ref{sec:sal_methods}) to gain insight into how the network operates and propagates protein configurations at a particular lag time.  

\begin{figure}[H]
    \centering
    \includegraphics[width=.8\textwidth]{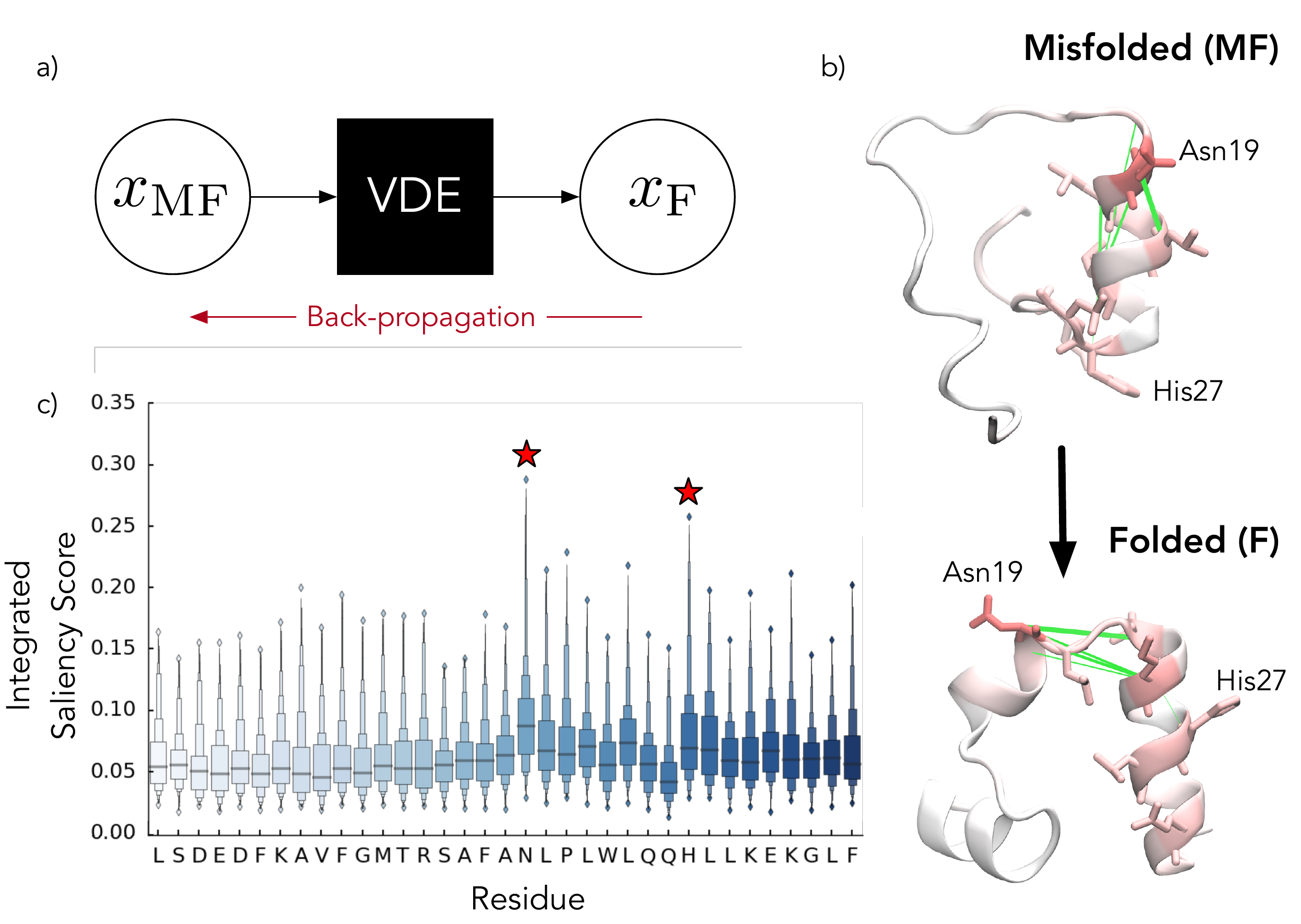}
    \caption{Protein saliency maps can be used to gain insight into the VDE. a) In saliency maps, the distances between the predicted and targeted output (i.e. contact distances in the folded states) are propagated back through the network to the input contact distances in order to gain insight into what the network learns. This is repeated for a large batch of possible configurations. b) For villin, the folded state is characterized by C$\alpha$ contact distances to the central Asparagine residue. In the misfolded state, this residue is too close to the first helix forming non-native contacts. The green lines denote the 5 contacts with the highest median saliency scores. c) Integrating over the saliency at the atomic level allows us to infer the importances of individual residues in certain state transitions, making them prime candidates (red stars) for further biophysical characterization. The distributions are computed over 200 transitions.}
    \label{fig:villin_sal}
\end{figure}
To perform the saliency analysis, we computed a median value for the derivative of the residual between villin's misfolded and folded basin with respect to its input contact features. As shown in Figure \ref{fig:villin_sal}b, these saliency maps for villin found 5 important C$\alpha$ contacts. The 5 contacts (indicated with green lines in Figure \ref{fig:villin_sal}b), all involve contacts to residues around Asn19. Remarkably, we can also integrate the saliency scores for each atomic feature to infer feature importances at the residue scale. The residue importance Figure \ref{fig:villin_sal}b-c can be used to potentially design new molecular simulations and biophysical experiments. For example, in the case of villin, our model predicts distances to Asn19 as being critical for movement out of the misfolded partially helical state (Figure \ref{fig:villin_sal}b). Mutating this residue to a proline or glycine could potentially be used to prevent the system from sampling the misfolded state. A potential drawback for this method is it requires sufficient knowledge of the system to identify a relevant path to investigate the corresponding initial and final conformations. This can be accomplished by either some empirical analysis, clustering, or simply sampling conformations at the minima and maxima of the latent space.

\section{Discussion}
In this work, we have introduced a variational autoencoder to analyze dynamical processes by incorporating a time lag into the traditional autoencoder structure, introducing an autocorrelation loss during training, and leveraging the Gaussian noise introduced into the latent space during training, dimensionality reduction, and synthetic trajectory analysis. Furthermore, we have introduced a saliency mapping approach inspired by advances in deep learning in order to interpret which features contribute to the identified reaction coordinate.

We demonstrate that the VDE is able to outperform state-of-the-art methods, such as tICA, in describing slow dynamics in both the 2-D M{\"u}ller-Brown potential and protein folding.
Using the 2-D system, we probe the generative nature of the VDE, showing that it can generate realistic thermodynamics and has some ability to extrapolate dynamics at temperatures it has not observed. Although this portion of our study is not quantitatively rigorous, we expect that a better understanding of how VDE parameters relate to simulation parameters might lead to using the VDE to cheaply generate reliable data for different simulation conditions and perhaps even protein mutants.
In the more complex case of protein folding, we show the utility of the VDE in understanding the conformational landscape of the villin headpiece domain, which is non-trivial due to the prominent misfolded state observed.
The latent space of the VDE captures the transitions among misfolded, unfolded, and folded, and a MSM constructed using the VDE projection exhibits a significantly longer timescale for the slowest process
than the optimized one constructed from tICA-transformed data. We also showed that we can peer into the ``black box'' that is the VDE via protein saliency mapping, which provides biophysical insights into the network's decision-making. For villin, we identify important C$\alpha$ contacts that we predict potentially play a role in misfolding-folding transitions. We anticipate that such results will prove useful in experimental design, such as in FRET experiments, to decide how to effectively probe a protein to observe conformational change.

While the VDE shows much promise, there are a few reasons why we cannot recommend it to completely replace previous methods, such as tICA, just yet. First, when training deep autoencoders using a autocorrelation loss (i.e.~as inspired by the VAC), there is a noticeable dependence on batch size that arises during training. The autocorrelation, as well as the related variational loss, attempts to calculate global equilibrium statistics, such as the exchange timescale for the slowest process. However, for finite batch sizes, we might only observe a single event in that process within a given batch. This may lead to underestimating the computed statistics since the network has no information about the rest of the dataset. This problem does not arise in tICA or MSMs because timescales and other global statistics are only estimated \textit{after} all the data has already been processed. Another issue with using the autocorrelation loss, as implemented, arises from the reality that many processes can occur with similar timescales. Each of these processes will be assigned highly similar autocorrelations, and thus might lead to volatile training; although, we believe our compound loss function can somewhat attenuate this issue, since the network is designed to keep track of global transition dynamics in addition to fitting to the slow processes.

All in all, VDEs and recent related work\cite{Wehmeyer2017, Mardt2017} herald exciting opportunities for bridging
Markov models and deep learning. We believe the expressive power of neural networks provides a natural solution to the choice-of-basis problem that plagues many Markovian analyses, while the strong theoretical underpinnings behind MSMs allow us to select and potentially even validate cross-validate neural architectures \cite{McGibbon2015, osprey}, ultimately allowing us to address fundamental questions in biophysics.

\section{Methods}

\subsection{M\"uller-Brown Potential}
\label{sec:muller_analysis}

We generated 10 independent simulations of the 2-D M\"uller-Brown potential governed by the following
equation:

\begin{equation*}
\dot{\mathbf{x}} = - \frac{\Delta V\left( \mathbf{x} \right)}{kT} + \sqrt{2 D} R\left( t \right),
\end{equation*}

\noindent{}where $k_{B}T = 1.5 \times 10^{4}$ Joules, $D = 10^{-2}$ meters-squared per second, and $R\left( t \right)$ is a
delta-correlated Gaussian process with zero mean, and $V\left(\mathbf{x}\right)$
is defined as:

\begin{equation*}
V\left(\mathbf{x}\right) = \sum^{4}_{j=1} A_j \cdot \exp \left[ a_j \left( x_1 - X_j \right)^2
+ b_j \left( x_1 - X_j \right) \left( x_2 - Y_j \right)
+ c_j \left( x_2 - Y_j \right)^2 \right],
\end{equation*}

\noindent{}where $a = \left(-1, -1, -6.5, 0.7 \right)$; $b = \left(0, 0, 11, 0.6\right)$;
$c = \left(-10, -10, -6.5, 0.7\right)$; $A = \left(-200, -100, -170, 15\right)$;
$X = \left(1, 0, -0.5, -1\right)$; and $Y = \left(0, 0.5, 1.5, 1\right)$ as
suggested by Müller and Brown~\cite{Muller1979}. Using the Euler-Maruyama method
for numerical integration and a time step of 0.1, we produced ten unique trajectories with $10^6$ time
steps, saved every 100 steps. The initial positions were sampled via a uniform
distribution over the box:
$\left[ -1.5, 1.2 \right] \times \left[ -0.2, 2.0 \right]$.

VDEs for the M\"uller-Brown potential were trained with a lag time of 10 time steps; 3 hidden layers with 256 nodes each; the Swish activation function \cite{ramachandran2017swish};
$\alpha$-value of $10^{-3}$; batch size of 100; dropout rate of 30\%; and a learning rate of $1 \times 10^{-4}$. We note that these parameters were not optimized using automated hyperparameter selection.
Gradient descent was performed with the Adam optimizer \cite{kingma2014adam}. Models were trained for 50 epochs, at which point the losses were observed to be converged. Prior to training, trajectories were preprocessed by subtracting their overall median values and scaling by inter-quartile ranges.

In constructing MSMs for the M\"uller-Brown potential, the scaled trajectories were then subject to dimensionality reduction using
principal component analysis (PCA)~\cite{Pearson1901}, time-structure based independent component analysis (tICA)~\cite{Naritomi2011, Perez-Hernandez2013, Schwantes2013},
and the pre-trained VDE, in order to generate one-dimensional representations of the system's dynamics.
We then partitioned the representations into twelves clusters using
the mini-batch $k$-means algorithm~\cite{Sculley2010, scikit-learn}.
Finally, the clusters were used to construct a maximum-likelihood estimated (MLE)
reversible Markov state model (MSM)~\cite{Metzner2009} with one timescale. A lag time of 10 time steps
was chosen for both MSM construction and dimensionality reduction, as the resulting
models provided optimal convergence of implied timescales. The MSMs were then evaluated 100 randomly seeded
hold-out datasets to generate unbiased GMRQ scores and standard errors.
All trajectory generation and analyses were performed with MSMBuilder~\cite{msmbuilder} and MSMExplorer~\cite{msmexplorer}.

Finally, in order to generate ``fake'' trajectories using the VDE, we randomly sampled initial positions via a uniform distribution, as described above,
and iteratively propagated these coordinates through the VDE for 1,000 steps, equivalent to 10,000 integrator steps. This was done for five scaling values, $\alpha$, evenly sampled in logspace between $10^{-2}$ and $10^{-1}$ to understand how the $\Lambda$-layer affects propagation.

\subsection{Villin Headpiece Domain}

We demonstrate the utility of the VDE method in characterizing the folding landscape of villin headpiece domain (pdb: 2f4k), a widely-studied 35-residue fast-folding protein, referred to henceforth as villin. Simulation data for villin was generated by Lindorff-Larsen et al.\cite{lindorff2011fast}. The simulation length was is $\mu$s and is strided at 2 ns for analysis.  C$\alpha$ contacts were used for featurization \cite{mdtraj, msmbuilder}.
VDEs for villin were trained with a lag time of 44 ns, selected to be the same as that in the optimal tICA model.
Other than expanding the number of hidden layers nodes to 1024, the training procedure was identical to that of Section \ref{sec:muller_analysis}.
The VDE was compared to an optimized tICA model for villin, as featurized with C$\alpha$ contacts, that was identified via hyperparameter optimization \cite{Husic2016}. Husic et al.\cite{Husic2016} have indicated that C$\alpha$ contacts are a useful featurization for representing folding processes \cite{Husic2016}, hence the selection of this featurization. In this model, a tICA lag time of 44 ns, 4 tICA components, and kinetic mapping \cite{noe2015kinetic} were selected according to hyperparameter optimization\cite{Husic2016}. 
To construct MSMs on tICA-transformed and VDE-transformed data, analogous steps as for the M\"uller potential in Section \ref{sec:muller_analysis} were performed. Mini-batch $k$-means clustering was performed with $125$ clusters for both sets of data. This was the optimal cluster number identified for tICA, and hyperparameter searching showed little influence of cluster number on MSMs from VDE-transformed data. MSMs for both tICA- and VDE- transformed models were constructed with 50 ns lag time. To obtain error estimates for MSM equilibrium populations and MSM timescales, 100 rounds of bootstrapping were performed over the original set of trajectories. The resulting ranges of values were used for error bars.

\subsection{Protein Saliency Maps}
\label{sec:sal_methods}
 Saliency maps \cite{Simonyan2013, Springenberg2014,Saliency2013} were originally proposed for looking at spatial support for varied classification problems. For image data, they find spatial features that a network looks for during classification, i.e., by asking how much does any individual pixel contribute to the final prediction. This is done by back-propagating from the desired class score through the network and into the image pixels. Similar to tICA loadings, the magnitude of the derivative can then be used to gauge feature importance per output class. An alternative closely related method, namely guided back-propagation, only propagates the positive derivatives through the network.
 
Saliency maps were designed for classification algorithms and thus needed to be modified for our application (Algorithm \ref{saliency}). Briefly, we first generate a faux two-step trajectory starting from a random protein conformation, for instance, a misfolded state, and going to the desired protein conformation, for instance, the folded state. The misfolded state is propagated through the network, and the residual to the folded state (Figure \ref{fig:villin_sal}a) is propagated back to obtain loadings on individual distances. Ideally, this is done for a large number of possible misfolded to folded transitions to obtain robust saliency maps. The median values for each feature across all of these maps can then be integrated to obtain residue level statistics or rank ordered to find important features.It is worth noting that  our method is different from classical saliency scoring whereby only the desired class label score is propagated backwards.

\begin{algorithm}[H]
\caption{Computing saliency maps}\label{saliency}
  \begin{algorithmic}[1]
    \Procedure{Saliency}{\texttt{model}, \texttt{data}}
    \State $x_t$, $x_{t+\tau} \gets$ \texttt{data}
    \State $x^\prime_{t+\tau} \gets$ \texttt{model.forward($x_t$)} 
        
    \State \texttt{model.loss} $= \norm{x^\prime_{t+\tau} - x_{t+\tau}}$
    \State \texttt{model.loss.backward()}
    \State \texttt{return} $\frac{\partial \texttt{model.loss}}{\partial x_t}$
    \EndProcedure
  \end{algorithmic}
\end{algorithm}

We note that both the VDE's noise parameter and the autocorrelation loss should be set to 0 for consistent results and numerical stability. We also recommend computing the saliency scores multiple times across many configurations and averaging out the results. Lastly, we note that the protein saliency maps can be used in a variety of different protein deep learning algorithms, including VAMPNets \cite{wu2017variational} and TAE \cite{Mardt2017}.

\section*{Availability}

Source code for this work is available under the open-source
MIT license and is accessible at \url{https://github.com/msmbuilder/vde}.
Complete examples can be found as Jupyter notebooks at
\url{https://github.com/msmbuilder/vde/tree/notebooks/}.

\section*{Author Contributions}
CXH, HKWS, MMS, and BEH performed analysis and wrote the manuscript.
CXH, HKWS, MMS, BEH, and VSP conceived of the methodology and edited the manuscript.
VSP supervised the project.

\section*{Acknowledgements}
We would like to thank J. Chodera, P. Eastman, E. Feinberg, and R. Sharma for insightful discussions.
We thank A. Peck for help copy-editing.
We acknowledge funding from NIH grants U19 AI109662 and 2R01GM062868 for their support of this work.
CXH and HKWS acknowledge support from NSF GRFP (DGE-114747). 
M.M.S would like to acknowledge support from the National Science Foundation grant NSF-MCB-0954714.
This work used the XStream computational resource, supported by the National Science Foundation Major Research Instrumentation program (ACI-1429830), as well as the Sherlock cluster, maintained by the Stanford Research Computing Center.

\section*{Disclosures}
VSP is a consultant and SAB member of Schrodinger, LLC and Globavir, sits on the
Board of Directors of Apeel Inc, Freenome Inc, Omada Health, Patient Ping,
Rigetti Computing, and is a General Partner at Andreessen Horowitz.

\bibliography{paper,annotated_vae_refs}

\end{document}